\documentclass[letterpaper]{article} % DO NOT CHANGE THIS
\usepackage{aaai23}  % DO NOT CHANGE THIS
\usepackage{times}  % DO NOT CHANGE THIS
\usepackage{helvet}  % DO NOT CHANGE THIS
\usepackage{courier}  % DO NOT CHANGE THIS
\usepackage[hyphens]{url}  % DO NOT CHANGE THIS
\usepackage{graphicx} % DO NOT CHANGE THIS
\usepackage{natbib}  % DO NOT CHANGE THIS AND DO NOT ADD ANY OPTIONS TO IT
\usepackage{caption} % DO NOT CHANGE THIS AND DO NOT ADD ANY OPTIONS TO IT
\usepackage{algorithm}
\usepackage{algorithmic}

\usepackage{tabularx}
\usepackage{bm} % make greek symbol bold
\usepackage{booktabs}
\usepackage{multirow}
\usepackage{amsmath}

\usepackage[hang,flushmargin]{footmisc}
\newcommand\workshopnote[1]{\renewcommand\thefootnote{}\footnote{#1}}

\usepackage{newfloat}
\usepackage{listings}
\DeclareCaptionStyle{ruled}{labelfont=normalfont,labelsep=colon,strut=off} % DO NOT CHANGE THIS
\floatstyle{ruled}
\newfloat{listing}{tb}{lst}{}
\floatname{listing}{Listing}
\nocopyright %-- Your paper will not be published if you use this command
\title{Can a Frozen Pretrained Language Model be used for Zero-shot Neural Retrieval on Entity-centric Questions?}
\author{
    Yasuto Hoshi,
    Daisuke Miyashita,
    Yasuhiro Morioka,
    Youyang Ng,
    Osamu Torii,
    Jun Deguchi
    }
\affiliations{
    Kioxia Corporation\\
    Kawasaki, Kanagawa, Japan\\
    \{yasuto1.hoshi, daisuke1.miyashita, yasuhiro.morioka, youyang.ng, osamu.torii, jun.deguchi\}@kioxia.com

}

\usepackage{bibentry}
\begin{document}

\maketitle

\begin{abstract}
	Neural document retrievers, including dense passage retrieval (DPR), have outperformed classical lexical-matching retrievers, such as BM25, when fine-tuned and tested on specific question-answering datasets.
	However, it has been shown that the existing dense retrievers do not generalize well not only out of domain but even in domain such as Wikipedia, especially when a named entity in a question is a dominant clue for retrieval.
	In this paper, we propose an approach toward in-domain generalization using the embeddings generated by the frozen language model trained with the entities in the domain.
	By not fine-tuning, we explore the possibility that the rich knowledge contained in a pretrained language model can be used for retrieval tasks.
	The proposed method outperforms conventional DPRs on entity-centric questions in Wikipedia domain and achieves almost comparable performance to BM25 and state-of-the-art SPAR model.
	We also show that the contextualized keys lead to strong improvements compared to BM25 when the entity names consist of common words.
	Our results demonstrate the feasibility of the zero-shot retrieval method for entity-centric questions of Wikipedia domain, where DPR has struggled to perform.
\end{abstract}

\workshopnote{Accepted to Workshop on Knowledge Augmented Methods for Natural Language Processing, in conjunction with AAAI 2023.}

\renewcommand{\thefootnote}{\arabic{footnote}}

% Neural document retrievers, including dense passage retrieval (DPR), have outperformed classical lexical-matching retrievers, such as BM25, when fine-tuned and tested on specific question-answering datasets. However, it has been shown that the existing dense retrievers do not generalize well not only out of domain but even in domain such as Wikipedia, especially when a named entity in a question is a dominant clue for retrieval. In this paper, we propose an approach toward in-domain generalization using the embeddings generated by the frozen language model trained with the entities in the domain. By not fine-tuning, we explore the possibility that the rich knowledge contained in a pretrained language model can be used for retrieval tasks. The proposed method outperforms conventional DPRs on entity-centric questions in Wikipedia domain and achieves almost comparable performance to BM25 and state-of-the-art SPAR model. We also show that the contextualized keys lead to strong improvements compared to BM25 when the entity names consist of common words. Our results demonstrate the feasibility of the zero-shot retrieval method for entity-centric questions of Wikipedia domain, where DPR has struggled to perform.

\section{Introduction}
Information retrieval (IR) is the task of finding relevant knowledge or passages corresponding to a given query.
Traditional exact lexical-matching approaches, such as TF-IDF (term frequency and inverse document frequency) or BM25 \citep{robertson1995okapi, robertson2009probabilistic}, have performed well on some IR tasks.
Recently, neural retrieval with contextualized dense sentence embedding has shown to be effective for tasks such as open-domain question answering \citep[e.g.,][]{karpukhin-etal-2020-dense,xiong2021approximate}.

Dense retriever, or dense passage retrieval (DPR), uses a pair of neural language models as a bi-encoder to obtain latent representations of questions and passages \citep{lee-etal-2019-latent}.
Bi-encoders require to be fine-tuned with contrastive learning to embed a question and the relevant passages on semantically similar sentence vectors.
Recent works on DPR reported in-domain (tested on the same dataset which was used to fine-tune) retrieval performances exceeding those from sparse retrievers such as BM25 \citep{karpukhin-etal-2020-dense}.

However, it has been shown that conventional DPRs struggle with retrieval on BEIR benchmark \citep{thakur2021beir} including various domains and even on EntityQuestions (EQ) \citep{sciavolino-etal-2021-simple} built within a confined domain (Wikipedia), whereas BM25 shows better performance on both of them.
This indicates that the generalizability of DPR is limited not only out of domain but even in domain when a named entity in a question is a dominant clue for retrieval.
In order to address this, \citet{sciavolino-etal-2021-simple} showed that DPR trained with PAQ dataset \citep{lewis-etal-2021-paq} which consists of 65 million of question--answer pairs containing many of named entities in Wikipedia still performs far worse than BM25 in EQ dataset.
The results showed that supervised learning with a huge dataset that covers the domain extensively could not lead to good generalization within the domain.

Though BM25 seems to be relatively robust, it may have an inherent drawback:
It is not good at searching for common terms in a query, even when the term is important, such as a part of a named entity.
Specifically, BM25 uses the inverse document frequency (IDF) weight of the query terms.
This property is based on the assumption that rare terms (of higher IDF values) are more informative than common terms (of lower IDF values).
However, there are not few cases where named entity consists of common words and a retriever must retrieve relevant passages using it as the dominant clue.
For instance, given a question ``\textit{Who is the author of Inside Job?}'', the entity name ``\textit{Inside Job}'' consists of common words, and the IDF values of these words are relatively low.
In this case, BM25 may have difficulty in retrieving relevant passages.

One of the reasons for the poor generalization of DPRs can be their poor retrieval accuracy on salient phrases such as named entities \citep[e.g.,][]{karpukhin-etal-2020-dense}.
To address the problem, \citet{Chen2021SalientPA} proposed the Salient Phrase Aware Retriever (SPAR) by combining existing DPR and a dense retriever which is trained so as to imitate BM25 prediction, and SPAR showed better performance than BM25 on EQ dataset.
Looking at SPAR from a perspective of modeling, it can be considered as the ensemble of two different dense retrievers.
In contrast, our motivation is to investigate if dense retrievers could generalize with a single language model without ensemble.

In this paper, we investigate whether a pretrained language model without supervised fine-tuning can be used for passage retrieval.
The approach is inspired by the idea that language models can have a lot more useful knowledge right after pretraining, but it has been forgotten during fine-tuning \citep[e.g.,][]{chen-etal-2020-recall}.
% By not fine-tuning the encoder model, we can employ the rich knowledge of the pretrained language model without overfitting to a certain dataset.
We propose \textit{Zero-shot Neural Retrieval} (Zero-NeR), a simple dense retrieval method that uses multiple contextualized keys for each passage generated from a frozen pretrained language model.
We show our zero-shot method outperforms conventional DPRs and achieves a retrieval accuracy comparable to that of BM25 and SPAR on entity-centric questions in Wikipedia domain.
Our system first extracts the named entities in questions and passages.
Then, a pretrained language model encodes the recognized entity names into contextualized queries and retrieval keys.

The key contributions of this paper are as follows:
\begin{enumerate}
	\item We propose a Zero-NeR method that can leverage rich representations for in-domain named entities from a pretrained and frozen language model, and show that its retrieval accuracy is better than the conventional DPRs.
	\item We show that using multiple retrieval keys for each passage improves recall especially when a named entity in a question is a dominant clue for retrieval.
	\item We demonstrate that our retrieval method is superior to BM25 when entity names in questions consist of common words.
\end{enumerate}

\section{Related Work}\label{sec:related}
\subsubsection{Passage Retrieval.}
Unsupervised sparse retrievers have traditionally been used for IR tasks, including answering open-domain questions \citep{chen-etal-2017-reading}.
They are based on bag-of-words exact lexical matching, including TF-IDF and a best-match weighting function called BM25 \citep{robertson1995okapi, robertson2009probabilistic}.
Unlike sparse retrieval, dense retrieval is based on semantic matching in the embedding space.
Dense retrievers leverage dense vector representations of sentences embedded by fine-tuned neural language models.
We refer the reader to \citet{thakur2021beir} for details on major retrieval methods.
It is also reported that re-ranking the retrieved passages can improve recall in EQ dataset \citep[e.g.,][]{sachan2022improving}, but note that re-rankers are outside the scope of this paper.

\subsubsection{In-Domain Generalization of Retriever.}
Dense retrievers still have room for improvement of in-domain generalization, as well as out-of-distribution generalization \citep{thakur2021beir}.
To generalize in Wikipedia domain, \citet{sciavolino-etal-2021-simple} trained DPR on PAQ dataset \citep{lewis-etal-2021-paq} which contains many of named entities in Wikipedia.
However, the trained model still performed far worse than BM25 in EQ dataset.
\citet{sciavolino-etal-2021-simple} also attempted to improve recalls for relational questions by training dedicated question encoders for each relation.
While this approach somewhat improved recall, the dedicated retrievers did not reach the performance of BM25 on average and did not solve the poor generalization problem.
In contrast, our proposed method employs a pretrained language model without fine-tuning for retrieval, allowing us to exploit the rich knowledge including named entities in Wikipedia learned by pretraining.

\subsubsection{Difference between Dense Retrieval and Sparse Retrieval.}
\citet{replication_dpr} showed that the overlap of results between dense and sparse retrieval was quite small.
It has been known empirically that the ensemble results for these relevance scores can exceed the performance of each alone \citep[e.g.,][]{karpukhin-etal-2020-dense}.
More recently, \citet{Chen2021SalientPA} attempted to train a dense retriever (called as dense Lexical Model $\bf{\Lambda}$) to imitate BM25 prediction.
Though $\bf{\Lambda}$ model underperforms BM25 by itself on EQ dataset, when combined with DPR trained on multiple QA datasets (DPR-multi), the combined model (SPAR) outperforms BM25.
SPAR requires two individually trained bi-encoders (thus four BERT models) with tangled architecture, whereas our motivation is to investigate if dense retrievers could generalize with a single language model without ensemble.
% In a more detailed analysis, \citet{sciavolino-etal-2021-simple} focused on the occurrence frequency in Wikipedia of the entity names in questions.
% They showed that the recall of the dense retriever varied greatly depending on the entity frequency, while the recall of the sparse retriever remained relatively constant regardless of the frequency.
We also investigate whether dense retrieval has any strengths in areas where sparse retrieval is lacking.
We experimentally demonstrate the differences between dense and sparse retrieval by using the IDF value to quantify the generality and rarity of an entity name in a question.

\begin{figure*}[t]
	\centering
	\includegraphics[width=1\textwidth]{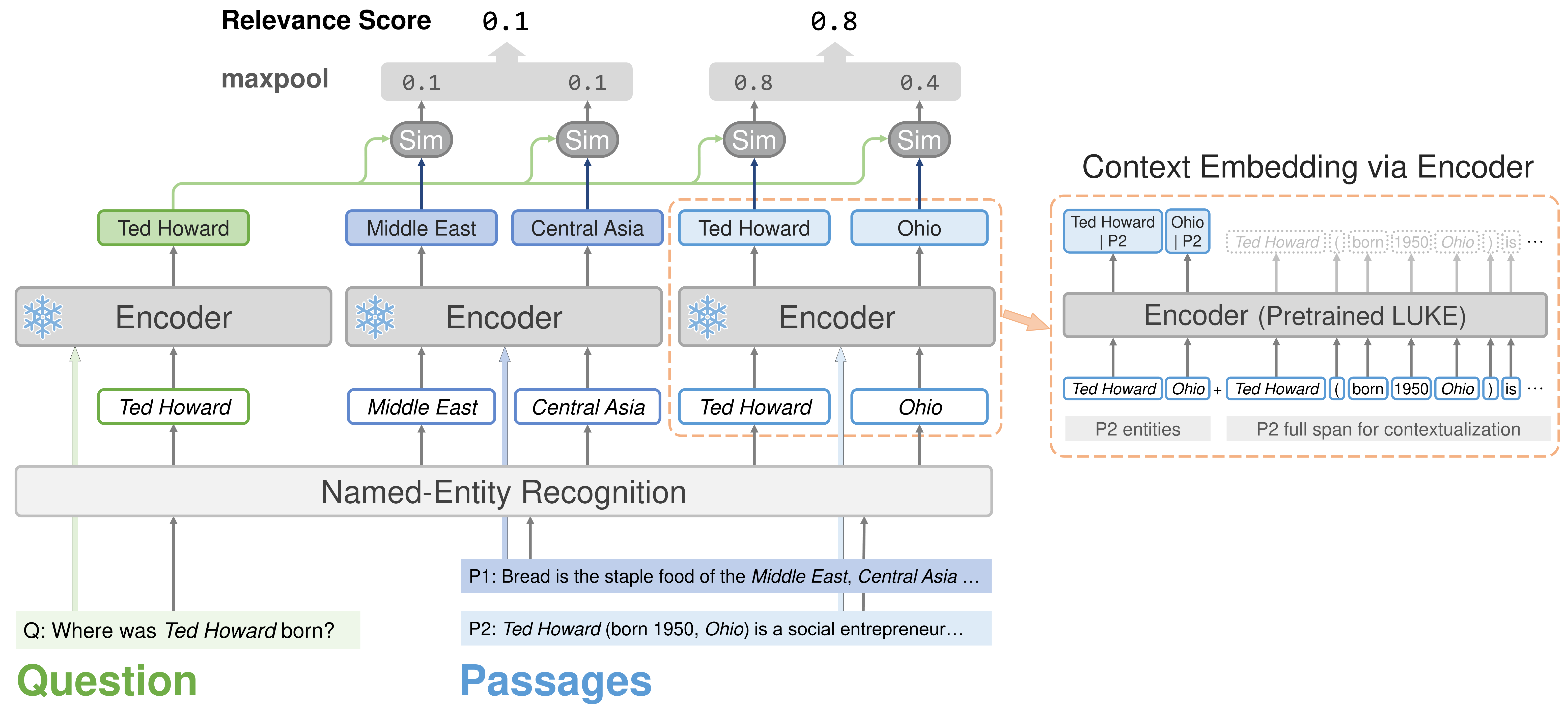}
	\caption{
		An overview of our proposed zero-shot dense retrieval system.
		Named entities in a question and passages are extracted via an off-the-shelf named-entity recognition model.
		Then, the extracted named entities (hollow rectangles) are encoded into dense representations (solid rectangles) with a frozen pretrained language model (denoted as \textit{Encoder}).
		Along with the entity span, a full sequence of the text is used to condition the semantics of the entire sentence on the embedding (outlined arrows).
		For instance, the dense representation corresponding to \textit{Ted Howard} is conditioned by P2, shown as ``Ted Howard $\mid$ P2'' in this Figure.
		We also use the embedding of the entire span of a passage title as a retrieval key.
		The passages can have multiple keys depending on how many entity names the passage has.
		The similarity between a query and a key is measured using cosine similarity (denoted as \textit{Sim}).
		For multiple keys in each passage, single relevance score for each passage is calculated via maximum pooling.
		Note that the proposed system requires no additional training if the encoder language model is pretrained.
	}
	\label{fig:overview}
\end{figure*}

\subsubsection{Multiple Keys for Passage Retrieval.}
Several methods have been proposed for calculating fine-grained interactions between a question and passages using multiple retrieval keys.
SPARTA (Sparse Transformer Matching, \citep{zhao-etal-2021-sparta}) and COIL (Contextualized Inverted List, \citep{gao-etal-2021-coil}) produce passage representations for each token embedded by fine-tuned language models and store them in the inverted index.
ColBERT \citep{khattab-etal-2021-relevance} leverages the embedding of all subword tokens in each passage to form a large collection of keys.
However, all of the methods listed here require the encoders for questions and passages to be fine-tuned.
During this fine-tuning for retrieval, multiple index updates are required, which is computationally expensive.
We seek a retrieval method that works in zero-shot setting, so we propose a method that only requires the inference operation of a pretrained language model together with named-entity recognition (NER).

\subsubsection{LUKE model} \citep{yamada-etal-2020-luke} is the state-of-the-art language model for tasks related to named entities, including NER.
% It also showed better performance for reading comprehension than the RoBERTa model on which it is based.
It is easy to suppose that the LUKE model, which seems to have useful knowledge about named entities, would also be useful for passage retrieval.
However, experiments using the model to perform passage retrieval have not been conducted.
In this paper, we explore the possibility that the rich knowledge contained in a pretrained language model can be used for retrieval tasks.
We show the advantage that our method has over existing dense retrievers for simple entity-centric questions.

\section{Zero-Shot Neural Retrieval}\label{sec:method}
\subsection{Pipeline}\label{sec:pipeline}
Here we describe our proposed zero-shot dense retrieval method using a pretrained and frozen language model.
Figure~\ref{fig:overview} provides a high-level overview of our system.
There are three major changes to the DPR: (1) employing embeddings of named entities in passages as retrieval keys, (2) scoring relevance with cosine similarity over multiple keys of each passage and maxpooling the score to output single score, and (3) elimination of encoder fine-tuning for retrieval.
The method follows the following steps:
\begin{enumerate}
	\item Named entities in questions and passages are extracted via the NER procedure.
	      Our implementation limits the number of named entities in a question to one.
	\item The named entities in questions and passages are encoded as a query and keys, respectively.
	      When encoding the named entities, the corresponding question and passages are input to the frozen language model to contextualize the named entities.
	      The title of the passage is also encoded and used as a key of the passage.
	      Therefore, a passage with $i$ entity names will have $i+1$ keys.
	\item The \textit{k} highest-scoring passages for a query are retrieved.
	      The cosine similarity is used as a scoring function.
	      For multiple keys in each passage, similarity scores to a query are calculated respectively, then single relevance score for each passage is calculated via maximum pooling.
	      Formally, given a query $q$, a total of $j$ keys $keys = (k_1, k_2, \dots, k_j)$ of a passage, and the similarity $s_i = \text{sim}(q, k_i)$ where $\text{sim}$ is cosine similarity, the relevance score for the passage is
	      \[
		      \text{score}(q, keys) = \text{maxpool}(s_1, s_2, \dots, s_j).
	      \]
\end{enumerate}

\subsection{Named Entity Recognition}\label{sec:method_ner}
We use the off-the-shelf LUKE model\footnote{\url{https://huggingface.co/studio-ousia/luke-large-finetuned-conll-2003}} \citep{yamada-etal-2020-luke} to extract named entities from passages.
This model is trained using entity-annotated Wikidata and is currently state-of-the-art in the CoNLL-2003 NER task \citep{tjong-kim-sang-de-meulder-2003-introduction}.
The inputs to the model are a text and the candidates of entity span in that text.
For each candidate span, the model predicts the types of entity (locations \texttt{LOC}, organizations \texttt{ORG}, persons \texttt{PER}, and miscellaneous \texttt{MISC}).

\begin{table*}[t]
	\centering
	\renewcommand{\arraystretch}{1.15}
	\begin{tabular}{llclc}
		\toprule
		\textbf{Modeling}       & \textbf{Retriever}                          & \textbf{\# LM} & \textbf{Retrieval training}          & \textbf{Recall@20 (macro average)} \\
		\midrule
		Sparse                  & BM25                                        & -              & -                                    & 72.0                               \\
		\hline
		\multirow{10}{*}{Dense} & DPR-NQ                                      & 2              & Supervised                           & 45.1                               \\
		                        & DPR-multi                                   & 2              & Supervised                           & 56.7                               \\
		                        & DPR-PAQ \citep{sciavolino-etal-2021-simple} & 2              & Supervised                           & 59.3                               \\
		                        & DPR-PAQ (SPAR checkpoint)                   & 2              & Supervised                           & 63.0                               \\
		                        & Wiki $\bf{\Lambda}$                         & 2              & Supervised                           & 69.0                               \\
		                        & PAQ $\bf{\Lambda}$                          & 2              & Supervised                           & 69.9                               \\
		                        & SPAR-Wiki (Wiki $\bf{\Lambda}$ + DPR-multi) & 4              & Supervised + Supervised              & 73.9                               \\
		                        & SPAR-PAQ (PAQ $\bf{\Lambda}$ + DPR-multi)   & 4              & Supervised + Supervised              & 74.5                               \\
		\cline{2-5}
		                        & Contriever                                  & 2              & Self-supervised                      & 64.7                               \\
		\cline{2-5}
		                        & \textbf{Zero-NeR (Ours)}                    & \textbf{1}     & \textbf{None (only LM pre-training)} & \textbf{67.1}                      \\
		\bottomrule
	\end{tabular}
	\caption{
		The macro average of the top-20 retrieval accuracies (denoted as Recall@20) on the 24 relations of EQ test set.
		\textit{\# LM} refers to the total number of language models used in each dense retriever with different weight parameters.
		\textit{Retrieval training} denotes fine-tuning methods for retrieval of each dense retriever.
		See text for details of the baseline models.
	}
	\label{tab:acc}
\end{table*}

\subsection{Contextualized Representations of Named Entities}\label{sec:contextualized_entity}
To produce queries and keys for retrieval, we use dense representations of the span-level embeddings of entity names in questions and passages.
We used the pretrained LUKE model\footnote{\url{https://huggingface.co/studio-ousia/luke-base}} to encode an entity span to a contextualized representation.
The LUKE model is based on the RoBERTa model \citep{roberta} and has an extension called \textit{entity-aware self-attention}.
This mechanism allows our system to embed a span-level contextualized representation of the entity name.
An entity embedding is output as a single vector from the model, even if it consists of multiple tokens.
We also use the embedding of the entire span of a title of a passage as a retrieval key.

\section{Experiments}\label{sec:experiments}
Here, we describe the datasets and baseline retrievers and explain the preparation for queries and retrieval keys with NER.
Then, we report the top-20 retrieval performances on EQ dataset.

\subsection{Datasets}\label{sec:datasets}
\textbf{Knowledge Source.}
We used the Wikipedia passage splits preprocessed by \citet{karpukhin-etal-2020-dense}.
The passages consist of English Wikipedia passages processed from Dec. 20, 2018 Wikipedia dump data.
The corpus contains a total of 3,232,908 articles containing 21,015,324 passages with a length of 100 words.

\textbf{EntityQuestions}\footnote{\url{https://github.com/princeton-nlp/EntityQuestions}} (EQ) consists of simple and entity-rich questions \citep{sciavolino-etal-2021-simple}.
The dataset contains questions from 24 common relations based on Wikidata \citep{vrandevcic2014wikidata}.
The questions were generated from relation triplets \textit{(subject, predicate, object)} in the T-REx dataset \citep{elsahar-etal-2018-rex} using 24 of manually defined question templates.
Each relation in the test set of EQ dataset has at most 1,000 questions (212 questions at minimum) and an average of 919.8 questions.
Because the number of questions differs by each relation, we follow the original paper and report the arithmetic macro-average of retrieval accuracies in each relation.

\subsection{Baseline Models}
\textbf{BM25} \citep{robertson2009probabilistic} is a bag-of-words retrieval function based on the term-matching of sparse vectors.
We used the Anserini \citep{anserini_Yang0L17} implementation with the default parameters of Lucene ($k = 0.9$ and $b = 0.4$).

DPR is a supervised method which employs a bi-encoder structure with a query encoder and a passage encoder \citep{karpukhin-etal-2020-dense}.
\textbf{DPR-NQ} is fine-tuned on the Natural Questions dataset \citep{kwiatkowski-etal-2019-natural}.
We used the model checkpoint provided in the DPR repo\footnote{\url{https://github.com/facebookresearch/DPR}}.
\textbf{DPR-multi} is fine-tuned on the combination of Natural Questions \citep{kwiatkowski-etal-2019-natural}, TriviaQA \citep{joshi-etal-2017-triviaqa}, WebQuestions \citep{berant-etal-2013-semantic}, and CuratedTREC \citep{curatedtrec}.
\textbf{DPR-PAQ} is fine-tuned on the PAQ dataset \citep{lewis-etal-2021-paq}.
In our evaluation, we used DPR-multi and DPR-PAQ checkpoints provided in the SPAR repo\footnote{\url{https://github.com/facebookresearch/dpr-scale}}.

\textbf{SPAR} \citep{Chen2021SalientPA} is a dense retriever that combines a standard DPR with the dense model called Lexical Model $\bf{\Lambda}$ trained to imitate BM25.
The SPAR model leverages two bi-encoders (thus four BERT models) to build itself and showed state-of-the-art performance on EQ dataset.
We use the \textbf{Wiki} $\bf{\Lambda}$ model trained on Wikipedia corpus, \textbf{PAQ} $\bf{\Lambda}$ model fine-tuned on PAQ dataset \citep{lewis-etal-2021-paq}, and models combined with the DPR-multi model, \textbf{SPAR-Wiki} (Wiki $\bf{\Lambda}$ + DPR-multi) and \textbf{SPAR-PAQ} (PAQ $\bf{\Lambda}$ + DPR-multi).
We used $\bf{\Lambda}$ and DPR-multi model checkpoints provided in the SPAR repo.
Following the original paper \citep{Chen2021SalientPA}, we tuned the concatenation weight $\mu$ of SPAR on development set of EQ dataset using the value of $\mu$ for which macro average of recall@100 is the best.
As a result of our tuning, $\mu = 1.25$ is the best for both SPAR-Wiki and SPAR-PAQ.

\textbf{Contriever}\footnote{\url{https://huggingface.co/facebook/contriever}} \citep{contriever} is trained with self-supervised contrastive learning on Wikipedia and CCNet \citep{wenzek-etal-2020-ccnet}.
To fine-tune the model with contrastive learning, a positive pair is formed by sampling two spans from a document, that is similar to Inverse Cloze Task \citep{lee-etal-2019-latent}.

% \textbf{Zero-shot RoBERTa.}\footnote{\url{https://huggingface.co/roberta-base}}
% RoBERTa model \citep{roberta} is the foundation of LUKE model which is used as the encoder in our method.
% For comparison with the proposed method that uses LUKE, token representations embedded by RoBERTa will be used as keys.
% We built the retrieval key for an entity span not as a single vector but as multiple embeddings of the corresponding subwords.
% For example, an entity span ``\textit{Ted Howard}'' is tokenized with [``\textit{Ted}'', ``\textit{Howard}''] by a subword tokenizer\footnote{Both the LUKE tokenizer and the RoBERTa tokenizer, from which this is based, tokenize a text into subwords through byte-level byte-pair encoding. The technique can handle rare words as corresponding subword units.}.
% This method of using subword tokens as retrieval keys is much like ColBERT \citep{khattab-etal-2021-relevance}, except for the following two points:
% (1) there is no fine-tuning for retrieval, and (2) only the tokens corresponding to the spans of named entities are used, not all tokens in the passage.

\subsection{Preparing Query and Retrieval Key with NER}\label{sec:q_k_gen}
\subsubsection{Query}
For the questions in EQ dataset, the named entity in a question can be extracted by using the corresponding question template.
For example, the question template for \texttt{\small place of birth} (P19) is ``\textit{Where was [E] born?}'', where \textit{[E]} is an entity name.
Given the question ``\textit{Where was Ted Howard born?}'', the entity name ``\textit{Ted Howard}'' can be extracted by comparing the question to the template.

\subsubsection{Retrieval Key}
For Wikipedia passages, following the NER method described in the previous section, a total of 219,258,978 named entities were recognized from 21-million passages.
Following the Pipeline section, the retrieval keys were made from named entities in the passages and the Wikipedia article titles.
Along with a total of 3,232,908 article titles, the total number of retrieval keys is 222,491,886.
This results in each passage having an average of 10.6 retrieval keys for our method.

% For preprocessing the keys of RoBERTa, the total number of keys was 602,980,362, which corresponding to subwords of named entity spans.

\subsection{Main Results: Retrieval Accuracy on the EQ Dataset}\label{sec:results_eq}
Table~\ref{tab:acc} shows the arithmetic macro average of the top-20 retrieval accuracies on the 24 relations of EQ test set.
Our method outperforms DPRs and Contriever and is almost comparable to BM25 and SPAR models.
When viewed with respect to each relation, our method outperforms BM25 in retrieval accuracy for seven relations and SPAR-PAQ for three relations.
Because the baseline DPRs employs asymmetric bi-encoder (query and key encoders do not share the parameters), \# LM of DPRs is two.
Note that, our method leverages the same model for the query and key encoders, thus \# LM of our method is one.

Contriever employs self-supervised learning method for retrieval training rather than supervised one where knowledge acquired during pretraining is prone to forget \citep[e.g.,][]{zhou-srikumar-2022-closer}.
However, Contriever performed below BM25 and our method, showing that self-supervised dense retrievers still have difficulties in answering entity-centric questions, even in-domain.

Overall, we find that our method outperforms conventional DPRs, while the performance is below BM25 and SPAR.
The results suggest that, by not fine-tuning, the knowledge of frozen language model acquired during pretraining can be applied to passage retrieval and could yield better in-domain generalization of dense retrieval.

% Note that our method outperformed DPR-NQ on all relations except \texttt{\small position played on team / speciality} (P413, \textit{What position does [E] play?}).

% For Zero-shot RoBERTa, the top-20 accuracy on EQ was 0.2\%.
% This result shows that even if multiple subwords are keyed for the named entity spans, the retrieval in the zero-shot setting may not work.
% This also suggests that there are advantages to embedding a named entity in a single contextualized representation.

\begin{figure}
	\centering
	\includegraphics[width=\linewidth]{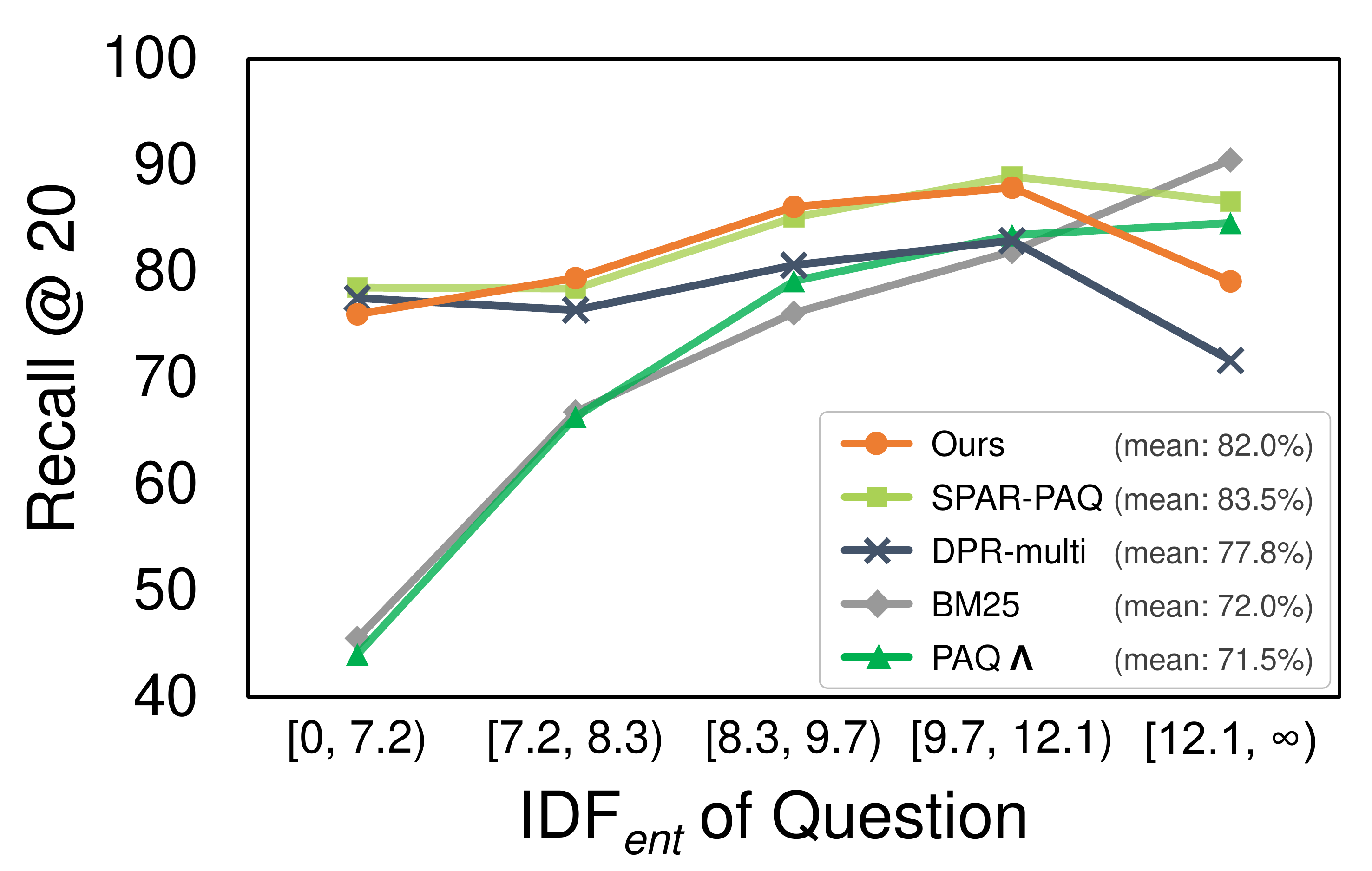}
	\caption{
		The top-20 retrieval accuracy on \texttt{\small author} (P50, \textit{Who is the author of [E]?}), where the test set questions were grouped by the $\mathrm{IDF}_{ent}$ value of the entity name in the questions.
		The questions are grouped into five buckets according to the maximum IDF value in the entity tokens.
		The number of questions in each bucket is 200.
		% The mean accuracies of BM25 and our method on the \texttt{\small author} questions are 73.0\% and 82.0\%, respectively.
	}
	\label{fig:idf}
\end{figure}

\begin{figure*}[t]
	\centering
	\includegraphics[width=0.85\textwidth]{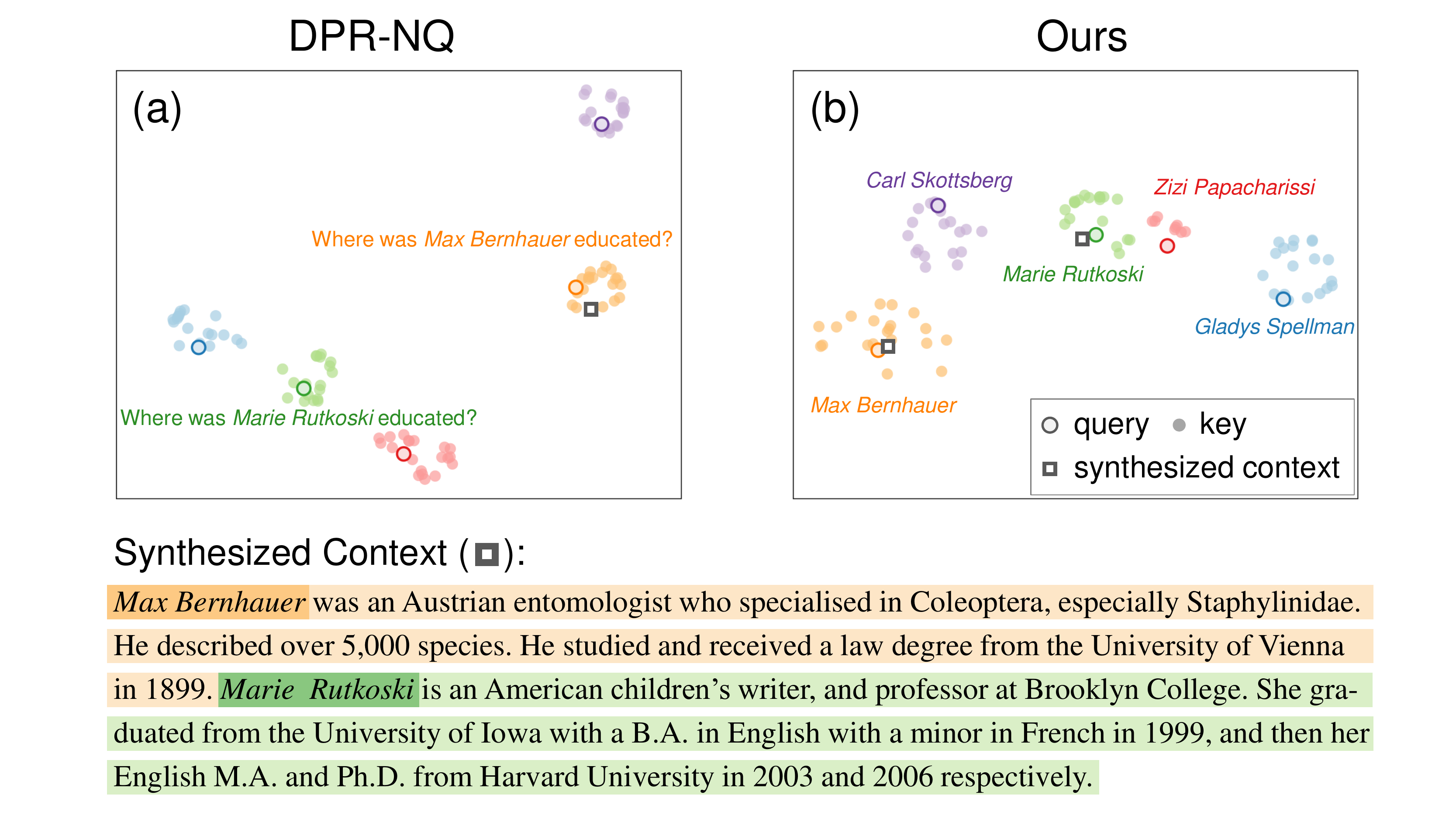}
	\caption{
		A UMAP projection of the queries and the corresponding top-20 keys retrieved by (a) DPR-NQ and (b) our method.
		For the queries, we used five randomly selected questions from the \texttt{\small educated at} questions (P69, \textit{Where was [E] educated?}).
		The retrieved keys with respect to a query (outlined circle) are plotted in the same color as the query.
		The outlined rectangles represent the embeddings of a synthesized context that contains multiple relational information points for two people, (\textit{Max Bernhauer} and \textit{Marie Rutkoski}).
		Note that the number of the plotted embeddings of the synthesized context is one for DPR-NQ but two for ours.
	}
	\label{fig:embedding_umap}
\end{figure*}

\section{Analysis}\label{sec:analysis}

\subsection{Characteristics of Our Method Compared with BM25}\label{sec:exp_idf}
Here, we examine the advantages our method has over BM25.
\subsubsection{Background}
Our method can have advantages in retrieval, especially when the named entities consist of common words.
Given a question like ``\textit{Who is the author of Inside Job?}'', it is expected that BM25 will have difficulty in searching relevant contexts.
This is because the entity name ``\textit{Inside Job}'' is composed of common words, and the IDF values of these words may be relatively low.
On the other hand, regardless of their IDF values, neural language models could distinguish that ``\textit{Inside Job}'' suggests the title of a work.
Such semantics could be useful information in dense retrieval.

\subsubsection{Methodology}
Specifically, we focus on the IDF value, which corresponds to the rarity of a word in a corpus.
Following the implementation of BM25, the IDF value of term $t$ is given by
$$ \mathrm{IDF}_t = \mathrm{log}\frac{N - n_t + 0.5}{n_t + 0.5} ,$$
where $N$ denotes the total number of passages and $n_t$ denotes the number of passages containing the term $t$.
For Wikipedia passages used in this study, the value of $N$ is 21,015,324 (see Datasets section).
We equally divided the \texttt{\small author} (P50, \textit{Who is the author of [E]?}) questions into five buckets according to the maximum IDF values of the entity words in them.
Each bucket has 200 questions.
We will henceforth refer to the maximum IDF value of the entity words as $\mathrm{IDF}_{ent}$.
For example, the IDF values of the words in ``\textit{Inside Job}'' (``\textit{Inside}'', ``\textit{Job}'') are 5.8 and 6.0, respectively.
Therefore, the $\mathrm{IDF}_{ent}$ value of ``\textit{Inside Job}'' is $\mathrm{IDF}_{ent} = \mathrm{max}(5.8, 6.0) = 6.0$.

\subsubsection{Result and Discussion}
Figure~\ref{fig:idf} shows the top-20 retrieval results as a function of $\mathrm{IDF}_{ent}$ of the named entity in the question.
Wiki $\bf{\Lambda}$ and SPAR-Wiki (not shown in the Figure) denote the similar tendency of PAQ $\bf{\Lambda}$ and SPAR-PAQ, respectively.

Not surprisingly, the retrieval accuracy of BM25 increases monotonically with respect to the $\mathrm{IDF}_{ent}$ values.
For lower $\mathrm{IDF}_{ent}$ values, the retrieval accuracy of BM25 was lower than that of our method.
Interestingly, PAQ $\bf{\Lambda}$ also shows a similar tendency.
This tendency is seemingly due to the fine-tuning for retrieval of $\bf{\Lambda}$ models where they learned to simulate the BM25 prediction.
This indicates that, especially for common words, simply learning to imitate BM25 is ineffective and contextualized representations of entities of our method are much more effective for retrieval.

DPR-multi, SPAR-PAQ, and our method perform with less bias for $\mathrm{IDF}_{ent}$ values.
Figure~\ref{fig:idf} indicates that the ensemble of PAQ $\bf{\Lambda}$ and DPR-multi achieves relatively flat $\mathrm{IDF}_{ent}$-dependence of SPAR and its high accuracies.
Our method achieves similar flatness despite using only one language model, whereas SPAR uses four and DPR-multi uses two different language models.

For the bucket with the largest $\mathrm{IDF}_{ent}$, the retrieval accuracy of our method and DPR-multi slightly decreased.
For our method, this result might reflect the nature of statistical language modeling: the model may have difficulty in giving meaningful representations for rare words because they have not appeared often enough for their representations to be learned \citep[e.g.,][]{match_your_words}.
For DPR-multi, the result suggests that the rarer the named entity in a question, the less it is included in the training dataset used to fine-tune, and the more difficult it is to retrieve in the test set.

\subsection{Single Key versus Multiple Keys: Comparison with DPR}\label{sec:comparison_with_dpr}
Here, we perform a detailed comparison of our method and DPR to assess the effect of using multiple retrieval keys.
\citet{luan-etal-2021-sparse} has shown the advantages of assigning multiple keys for each passage via so-called multi-vector encoding.
Our method, likewise, should have an advantage because it uses multiple keys, especially when the passage has multiple relational information.

\subsubsection{Result and Discussion}
Figure~\ref{fig:embedding_umap} shows a UMAP projection \citep{umap} of the queries and the corresponding keys retrieved by DPR-NQ and our method.
The queries consist of five randomly selected questions of the type \texttt{\small educated at} (P69, \textit{Where was [E] educated?}).
For DPR-NQ, the query was made from a full sequence of a given question.
The query for our method was the embedding of an entity name (in this case, the name of a person) extracted from the question, where the corresponding question was also input to the encoder to contextualize the entity (see Pipeline section for details).
The keys in Figure~\ref{fig:embedding_umap} indicate the top-20 retrieval results for DPR-NQ and our proposed method.
In addition to the Wikipedia passages, we synthesized a passage that contains multiple relational information for two people, (\textit{Max Bernhauer} and \textit{Marie Rutkoski}).

Note that DPR produces a single key from the synthesized context, whereas our method produces multiple keys.
Our method successfully retrieved the synthesized context with both queries (\textit{Max} and \textit{Marie}), but DPR-NQ failed to retrieve it with the query \textit{Marie}.
Thus, the multiple keys of our method can help retrieve a passage when it contains multiple relational information.
In the next section, we report a detailed ablation that uses one key for each passage selected from the multiple keys.

\begin{figure}
	\centering
	\includegraphics[width=\linewidth]{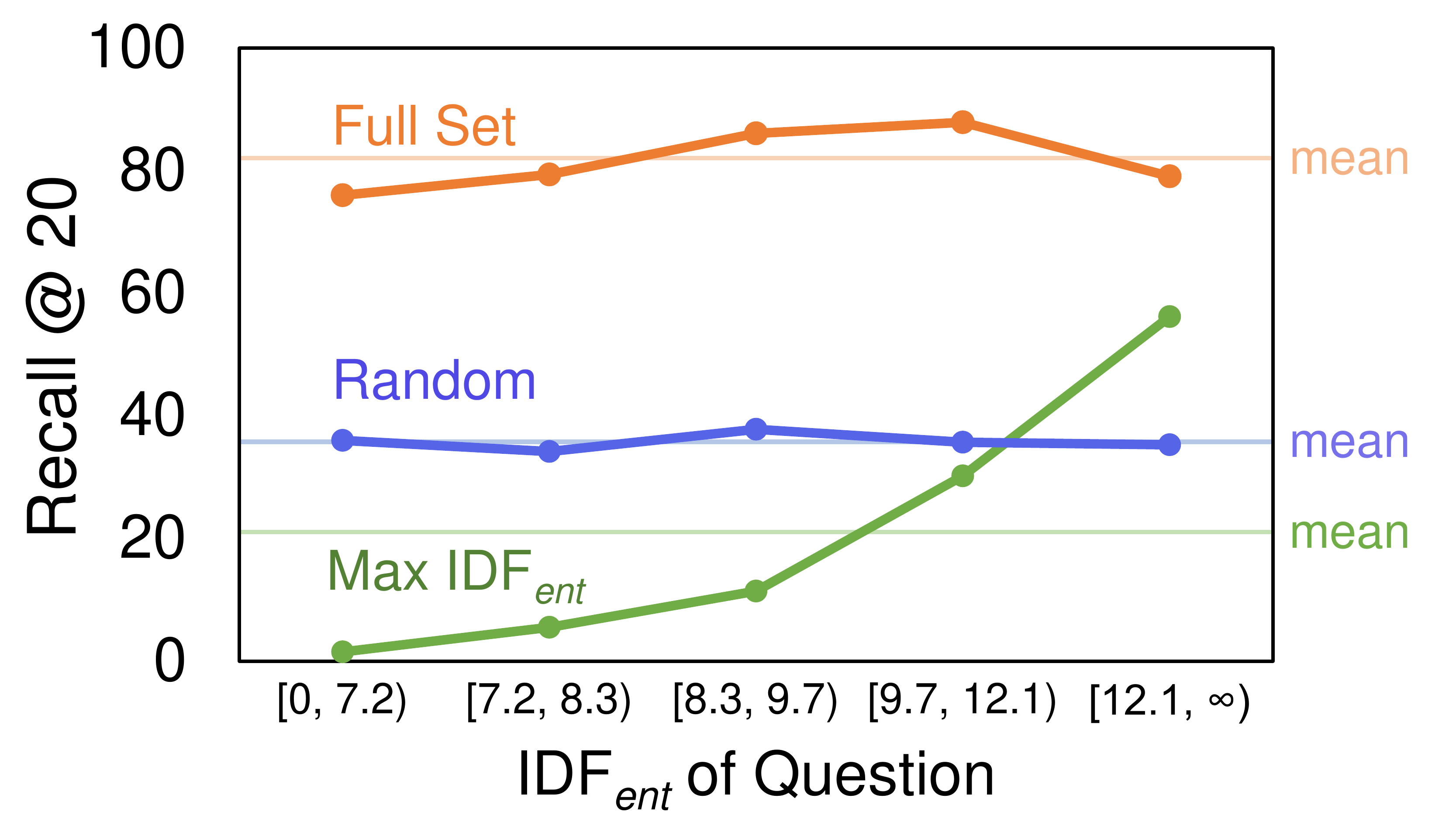}
	\caption{
		The top-20 retrieval accuracy for \texttt{\small author} (P50, \textit{Who is the author of [E]?}).
		The format is the same as in Figure~\ref{fig:idf},
		and the Full Set is the same result as the one in that figure.
		For each passage from Wikipedia, only one key was selected using random selection (denoted as Random), along with the key corresponding to the maximum $\mathrm{IDF}_{ent}$ (denoted as Max $\mathrm{IDF}_{ent}$).
		The mean accuracies of Random and Max $\mathrm{IDF}_{ent}$ are 35.8\% and 21.0\%, respectively, while the mean accuracy of Full Set is 82.0\%.
	}
	\label{fig:one_key}
\end{figure}

\subsection{Ablation Study: Are Multiple Retrieval Keys Required?}\label{sec:ablation_multiple_keys}
Here, we report the retrieval results from using one key for each passage, an equivalent setting to conventional dense retrieval, so that we may better assess the advantages that we have by using multiple keys.
As described in Experiments section, the average number of retrieval keys was 10.6 for each 100-word passage from Wikipedia.
We chose only one key from the full set of multiple keys for each passage in the following two sampling forms:
(1) selected the key randomly (referred to as Random) and (2) selected a key corresponding to an entity span of maximum $\mathrm{IDF}_{ent}$ (referred to as Max $\mathrm{IDF}_{ent}$).

\subsubsection{Results and Discussion}
In Figure~\ref{fig:one_key}, the top-20 retrieval results are shown with the keys sampled using the forms above.
The result with a full set of keys clearly surpasses the results with a single key (Random and Max $\mathrm{IDF}_{ent}$).
Not surprisingly, the result for Random is flat with no sensitivity to the $\mathrm{IDF}_{ent}$ value.
However, for Max $\mathrm{IDF}_{ent}$, the retrieval accuracy decreases as the $\mathrm{IDF}_{ent}$ value decreases.
This could be because the corresponding key could not be registered when the $\mathrm{IDF}_{ent}$ value for a question was small.
For instance, the passages corresponding to the \texttt{\small author} questions can have the title of the work and the author's name.
If $\mathrm{IDF}_{ent}$ is small for a question (such as ``\textit{Inside Job}'', the title of the work), then it is easy to imagine that the author's name in the same passage could have a larger $\mathrm{IDF}_{ent}$.
In this case, the retrieval accuracy can drop off in the absence of keys relevant to the question.
We should also mention that even for larger $\mathrm{IDF}_{ent}$, the retrieval accuracy of Max $\mathrm{IDF}_{ent}$ is below that of a full set of keys.
To summarize, we argue that using multiple keys for each passage significantly impacts on passage retrieval.

\subsection{Ablation Study: Is ``Contextualized'' Embedding of ``Named Entity'' Required?}\label{sec:ablation_contextualized_embedding}
In this section, we investigate the impact that the input to the frozen LUKE encoder model during query generation has on retrieval accuracy.
% As described in Experiment section, zero-shot RoBERTa on EQ dataset achieved 0.2\% of recall.
% The result suggests that multiple keys for subwords of named entity are not practical for zero-shot retrieval and that there may be advantages to embedding named entities in a single contextualized representation.

\subsubsection{Background}
For the entity-aware self-attention mechanism in the LUKE model, a full sentence is input along with entity spans as a condition of the entity spans.
The mechanism considers the context for the entity name when computing attention scores.
For instance, given the question ``\textit{Where was Ted Howard born?}'', the contextualized representation of the entity name ``\textit{Ted Howard}'' is calculated under the condition of a question about the birthplace.
We focused on this conditioning and examined the effect of contextualization on queries.

\subsubsection{Result and Discussion}
Table~\ref{tab:conditioning} shows the effectiveness of contextualization.
When the query embeddings were generated with only entity spans, the retrieval accuracy degraded about 4 points.
Furthermore, the accuracy significantly decreased when the query embeddings were generated for the entire question spans.
This indicates that it is necessary to use the contextualized embeddings of named entities in questions.

\begin{table}[t]
	\centering
	\begin{tabular}{llcc}
		\toprule
		\textbf{Query} & \textbf{Condition} & \textbf{Top-20} & \textbf{Top-100} \\
		\midrule
		Entity span    & Full span          & \textbf{67.1}   & \textbf{77.5}    \\
		Entity span    & Entity span        & 63.2            & 73.6             \\
		Full span      & Full span          & 28.6            & 44.7             \\
		\bottomrule
	\end{tabular}
	\caption{
		The top-20 and top-100 retrieval accuracies (\%) on EQ test sets. The highest score is indicated in bold.
		For example, given a question ``\textit{Where was Ted Howard born?}'', the entity span is ``\textit{Ted Howard}'', and the full span of questions is ``\textit{Where was Ted Howard born?}''.
	}
	\label{tab:conditioning}
\end{table}

\section{Conclusion}
In this paper, we investigate whether a pretrained language model without supervised fine-tuning can be used for passage retrieval.
We proposed \textit{Zero-NeR}, a simple zero-shot neural retrieval method employing rich knowledge in a pretrained and frozen language model for retrieval.
Toward in-domain generalization, we investigate the feasibility of the proposed method for entity-centric questions in Wikipedia domain, where existing dense retrievers has struggled to perform.
Our method outperforms the conventional supervised and self-supervised bi-encoders, and it is almost comparable to the lexical-matching model and the state-of-the-art dense retriever.
We also show that the contextualized keys lead to strong improvements compared to BM25 when the entity names consist of common words.
Our findings suggest that zero-shot representations from the frozen language model can be used for retrieval on entity-centric questions and also lead better in-domain generalization.

\bibliography{retriever_zeroner}

% \clearpage

\appendix
\section{Appendix}\label{sec:appendix}

\subsection{Full Results on EntityQuestions Dataset}\label{sec:appendix_full_results}
Table~\ref{tab:resutls_eq} shows the evaluation results on EQ test set.
Our proposed Zero-NeR outperformed both DPRs and Contriever on average.
Also, our method does not show extreme performance degradation for relations like P19 or P106 that have very low performance for DPR-NQ.
Note that our retriever has not been fine-tuned at all, whereas the other dense retrievers have been.

\begin{table*}
	\centering
	% \small
	% {\scriptsize
	\scalebox{0.65}{
		\begin{tabular}{llcccccccccc}
			\toprule
			Relations                                             & Question templates                      & \textbf{BM25}    & \textbf{DPR-NQ} & \textbf{DPR-multi} & \textbf{DPR-PAQ} & \textbf{Wiki} $\bf{\Lambda}$ & \textbf{PAQ} $\bf{\Lambda}$ & \textbf{SPAR-Wiki} & \textbf{SPAR-PAQ} & \textbf{Contriever} & \textbf{Zero-NeR (Ours)} \\ %\multicolumn{2}{l}{Relations}
			\midrule

			P17                                                   & Which country is [E] located in?        & 61.5             & 56.6            & 67.7               & 68.4             & 66.2                         & 65.0                        & \underline{70.2}   & \textbf{70.6}     & 65.4                & 65.6                     \\
			P19                                                   & Where was [E] born?                     & \textbf{75.3}    & 26.0            & 41.6               & 65.0             & 73.6                         & 73.4                        & 73.3               & \underline{73.9}  & 63.0                & 71.6                     \\
			P20                                                   & Where did [E] die?                      & \textbf{80.4}    & 32.8            & 45.1               & 67.2             & 76.3                         & 78.2                        & \underline{78.9}   & \textbf{80.4}     & 70.6                & 69.3                     \\
			P26                                                   & Who is [E] married to?                  & \textbf{89.7}    & 25.1            & 48.2               & 66.5             & 83.0                         & 83.2                        & 84.8               & \underline{85.0}  & 73.2                & 63.6                     \\
			P36                                                   & What is the capital of [E]?             & 90.6             & 74.9            & 78.8               & 82.5             & \underline{91.4}             & 88.6                        & \textbf{92.6}      & \underline{91.4}  & 89.7                & 81.8                     \\
			P40                                                   & Who is [E]'s child?                     & \textbf{85.0}    & 16.5            & 33.7               & 57.8             & 81.3                         & 81.9                        & 81.1               & \underline{83.2}  & 66.4                & 52.5                     \\
			P50                                                   & Who is the author of [E]?               & 73.0             & 75.7            & 77.8               & \textbf{84.9}    & 69.0                         & 71.5                        & 82.7               & \underline{83.5}  & 74.6                & 82.0                     \\
			P69                                                   & Where was [E] educated?                 & 73.1             & 19.9            & 41.9               & 59.3             & 70.4                         & 73.1                        & \underline{73.7}   & \textbf{74.9}     & 56.0                & 61.1                     \\
			P106                                                  & What kind of work does [E] do?          & 71.2             & 19.9            & 53.0               & 48.5             & 65.5                         & 68.9                        & 77.0               & \textbf{79.8}     & 66.2                & \underline{79.6}         \\
			P112                                                  & Who founded [E]?                        & 81.2             & 74.7            & 75.7               & 77.3             & 77.6                         & 78.8                        & \textbf{84.3}      & \underline{84.1}  & 77.1                & 74.7                     \\
			P127                                                  & Who owns [E]?                           & \underline{78.4} & 46.5            & 63.8               & 65.6             & 74.5                         & 75.6                        & 77.1               & \textbf{78.5}     & 69.1                & 64.9                     \\
			P131                                                  & Where is [E] located?                   & \textbf{63.1}    & 44.1            & 44.1               & 39.1             & \underline{58.0}             & 57.0                        & 57.3               & 56.2              & 52.5                & 50.4                     \\
			P136                                                  & What type of music does [E] play?       & 48.7             & 34.7            & 36.9               & 48.4             & 45.0                         & 48.2                        & 51.5               & \underline{54.4}  & 45.6                & \textbf{57.0}            \\
			P159                                                  & Where is the headquarter of [E]?        & 85.0             & 69.0            & 72.0               & 82.1             & 84.0                         & 84.1                        & \underline{86.4}   & \textbf{87.1}     & 79.7                & 78.1                     \\
			P170                                                  & Who was [E] created by?                 & 72.6             & 33.4            & 57.7               & 67.7             & 68.9                         & 68.7                        & \underline{75.7}   & \textbf{77.5}     & 59.7                & 72.3                     \\
			P175                                                  & Who performed [E]?                      & 56.6             & 41.6            & 51.6               & 41.8             & 49.8                         & 52.4                        & 60.9               & \underline{	61.7 	} & 47.5                & \textbf{63.9}            \\
			P176                                                  & Which company is [E] produced by?       & 81.0             & 43.0            & 73.7               & 69.6             & 79.4                         & 80.2                        & \underline{81.9}   & \textbf{82.0}     & 71.7                & 74.5                     \\
			P264                                                  & What music label is [E] represented by? & 45.6             & 27.6            & 43.0               & \underline{55.0} & 42.0                         & 44.8                        & \textbf{56.2}      & \underline{55.0}  & 37.2                & 50.4                     \\
			P276                                                  & Where is [E] located?                   & 84.9             & 71.4            & 77.3               & 76.1             & 83.1                         & 82.8                        & \textbf{85.6}      & \underline{85.3}  & 78.3                & 77.8                     \\
			P407                                                  & Which language was [E] written in?      & 86.2             & 72.9            & 82.5               & 85.9             & 83.6                         & 85.5                        & \underline{87.5}   & \textbf{88.2}     & 81.7                & 79.1                     \\
			P413                                                  & What position does [E] play?            & 74.3             & 75.7            & 71.4               & 54.4             & 71.2                         & 73.5                        & \textbf{82.7}      & \underline{82.2}  & 69.3                & 73.0                     \\
			P495                                                  & Which country was [E] created in?       & 21.8             & 19.4            & \textbf{27.9}      & \underline{26.3} & 20.9                         & 21.3                        & \underline{26.3}   & 26.1              & 21.5                & 21.5                     \\
			P740                                                  & Where was [E] founded?                  & 74.4             & 57.0            & 61.6               & 78.0             & 69.5                         & 71.1                        & \textbf{78.8}      & \underline{78.3}  & 73.1                & 77.5                     \\
			P800                                                  & What is [E] famous for?                 & \textbf{74.7}    & 24.4            & 34.4               & 45.3             & \underline{71.5}             & 69.2                        & 67.9               & 67.4              & 64.7                & 69.2                     \\

			\midrule
			\multicolumn{2}{l}{\textit{Arithmetic macro average}} & 72.0                                    & 45.1             & 56.7            & 63.0               & 69.0             & 69.9                         & \underline{73.9}            & \textbf{74.5}      & 64.7              & 67.1                                           \\
			\multicolumn{2}{l}{\textit{Arithmetic micro average}} & 71.4                                    & 44.6             & 56.5            & 62.8               & 68.4             & 69.4                         & \underline{73.6}            & \textbf{74.1}      & 64.1              & 66.6                                           \\
			\multicolumn{2}{l}{\textit{Best on}}                  & 6                                       & 0                & 1               & 1                  & 0                & 0                            & 6                           & 9                  & 0                 & 2                                              \\
			\bottomrule
		\end{tabular}
	}
	\caption{
		The top-20 retrieval accuracy on the test sets of EntityQuestions dataset. The highest score is indicated in bold, and the second highest score is underlined.
	}
	\label{tab:resutls_eq}
\end{table*}

\end{document}